\documentclass{article}

\newif\ifpdf
\ifx\pdfoutput\undefined
  \pdffalse
\else
  \pdfoutput=1
  \pdftrue
\fi

\ifpdf
  \usepackage[pdftex]{graphicx}
  \usepackage[pdftex]{color}
  \DeclareGraphicsExtensions{.pdf,.png,.jpg}
\else
  \usepackage[dvips]{graphicx}
  \usepackage[dvips]{color}
  \DeclareGraphicsExtensions{.eps,.epsi,.ps}
\fi

\usepackage{times}
\usepackage{amsfonts}
\usepackage{latexsym}
\usepackage{fullpage}

\title{Qualitative Analysis of Correspondence \\ for Experimental Algorithmics}

\author{
Chris Bailey-Kellogg \\
1398 Computer Science Building\\
Department of Computer Sciences \\
Purdue University, IN 47907\\
{\tt cbk@cs.purdue.edu}
\and
Naren Ramakrishnan \\
629 McBryde Hall \\
Department of Computer Science \\
Virginia Tech, VA 24061\\
{\tt naren@cs.vt.edu}
}

\newlength\colwidth 

\long\def\gobble#1{}
\def\set#1{{\{#1\}}}
\def\R{{\ensuremath\mathbb{R}}}

\newenvironment{closeitemize}{\begin{list}{-}{\topsep=0in\itemsep=0in\parsep=0in}}{\end{list}}

\begin{document}

\maketitle

\begin{abstract}

Correspondence identifies relationships among objects via similarities
among their components; it is ubiquitous in the analysis of spatial
datasets, including images, weather maps, and computational
simulations.  This paper develops a novel multi-level mechanism for
qualitative analysis of correspondence.  Operators leverage domain
knowledge to establish correspondence, evaluate implications for model
selection, and leverage identified weaknesses to focus additional data
collection.  The utility of the mechanism is demonstrated in two
applications from experimental algorithmics --- matrix spectral
portrait analysis and graphical assessment of Jordan forms of
matrices.  Results show that the mechanism efficiently samples
computational experiments and successfully uncovers high-level problem
properties.  It overcomes noise and data sparsity by leveraging domain
knowledge to detect mutually reinforcing interpretations of spatial
data.

\end{abstract}

\section{Introduction}

\begin{figure}
\begin{center}
\includegraphics[width=3in]{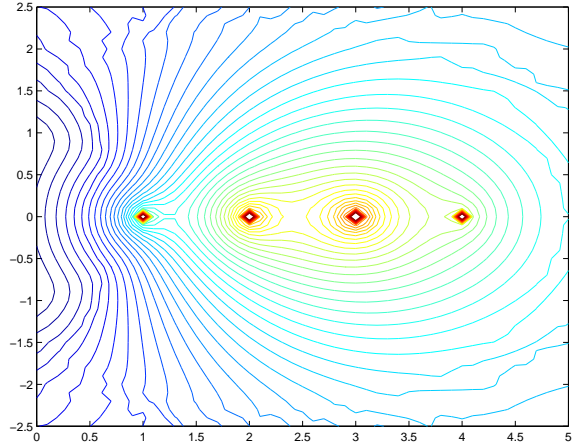} 
\end{center}
\caption{An example spectral portrait, for a matrix with eigenvalues
at 1, 2, 3, and 4.  Qualitative properties of the portait correspond
to important characteristics of the underlying problem that must be
considered with respect to algorithm selection. For instance, a level
curve surrounding multiple eigenvalues indicates a level of precision
beyond which the eigenvalues cannot be distinguished.}
\label{fig:compan}
\end{figure}

Correspondence is a ubiquitous concept in the interpretation of
spatial datasets.  Correspondence establishes analogy, indicating
objects that play similar roles with respect to some context.  For
example, correspondence between template and image features supports
object recognition, correspondence among isobars in a weather map aids
identification of pressure troughs and ridges, and, as shown in this
paper, correspondence and lack thereof among level curves in datasets
like Fig.~\ref{fig:compan} supports characterization of matrix
properties for scientific computing applications.

This paper develops a novel qualitative analysis mechanism for
extracting and utilizing correspondence in spatial data, with
particular focus on model selection in applications with sparse, noisy
data.  We develop our mechanism in the context of {\em experimental
algorithmics}, an emerging field that designs methodologies for
empirically evaluating algorithms on realistic test problems, and
interpreting and generalizing the results to guide selection of
appropriate mathematical software.  It is the preferred method of
analysis in applications where domain knowledge is imperfect and for
which our understanding of the factors influencing algorithm
applicability is incomplete.  For instance, when solving linear
systems associated with finite-difference discretization of elliptic
partial differential equations (PDEs), there is little mathematical
theory to guide a choice between, say, a direct solver and an
iterative Krylov solver plus preconditioner.  An experimental
algorithmics approach is to parameterize a suitable family of
problems, and mine a database of PDE ``solves'' to gain insight into
the likely relative performance of these two
approaches~\cite{naren-ribbens}.  Thus experimental algorithmics is
data-driven and aims to empirically capture the relationship between
problem characteristics and algorithm performance.

Many tasks in experimental algorithmics involve assessing the
eigenstructure of a given matrix.  Eigenstructure helps characterize
the stability, sensitivity, and accuracy of numerical methods as well
as the fundamental tractability of problems.  Recently, the {\em
spectral portrait} (e.g.\ Fig.~\ref{fig:compan}) has emerged as a tool
for graphically visualizing eigenstructure.  A spectral portait
characterizes how the eigenvalues of a matrix change as perturbations
(e.g.\ due to numerical error) are introduced in computations
involving the matrix.  Level curves in a spectral portait correspond
to perturbations, and the region enclosed by a level curve contains
all possible eigenvalues that are equivalent with respect to
perturbations of a given magnitude.  Analysis of level curves with
respect to a class of perturbations reveals information about the
matrix (e.g.\ nonnormality and defective eigenvalues) and the effects
of different algorithms and numerical approximations.

Experimental algorithmics thus determines {\em high-level properties}
from analysis of data from {\em low-level computational experiments};
we focus here on the particular case where such properties are
extracted from {\em graphical representations} and where it is
necessary to minimize the computational experiments performed (owing
to the cost and complexity of conducting them).  We pose the
extraction of high-level properties as a model selection problem and
show that correspondence can be exploited to drive data collection in
order to discriminate among possible models.  Each step in our
framework is parameterized by domain knowledge of properties such as
locality, similarity, and correspondence.  Therefore the
framework is generic both with respect to a variety of problems in
experimental algorithmics (two case studies are presented here), and
to problems in other domains (e.g.\ connections to weather data
analysis are discussed).  Furthermore, the approach leads to efficient
and explainable data collection motivated directly by the need to
disambiguate among high-level models.

\section{Qualitative Analysis of Spatial Data}

\begin{figure}
\begin{center}
\includegraphics[width=3in]{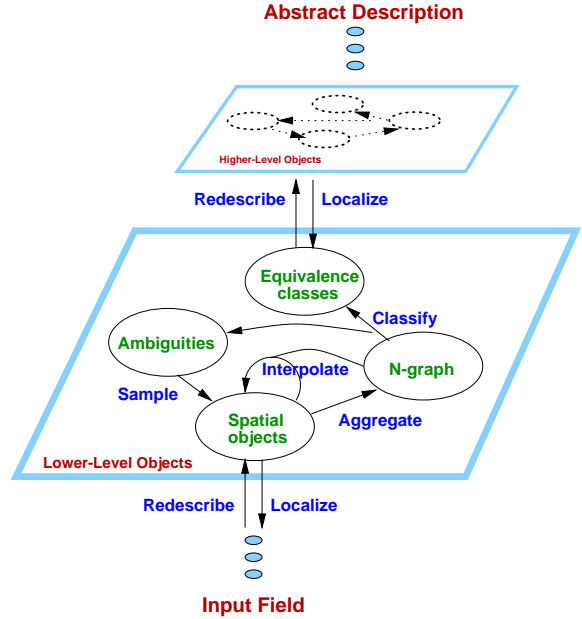}
\end{center}
\caption{The spatial aggregation language provides a uniform vocabulary of 
operators utilizing domain knowledge to build multi-layer structural
descriptions of spatial data.}
\label{fig:sal}
\end{figure}

Our mechanism for qualitative analysis of correspondence is based on
the Spatial Aggregation Language (SAL)~\cite{bailey-kellogg96,yip96a}
and the ambiguity-directed sampling
framework~\cite{bailey-kellogg-naren01}.  SAL programs apply a set of
uniform operators and data types (Fig.~\ref{fig:sal}) in order to
extract multi-layer geometric and topological representations of
spatial data.  These operators utilize domain knowledge of physical
properties such as continuity and locality, specified as metrics,
adjacency relations, and equivalence predicates, to uncover regions of
uniformity in spatially distributed data.  Ambiguity-directed sampling
focuses data collection so as to clarify difficult choice points in an
aggregation hierarchy.  It seeks to maximize information content while
minimizing the number and expense of data samples.

As an example of aggregation, consider the construction of level
curves in the spectral portrait in Fig.~\ref{fig:compan}.  The key
steps in such an analysis (after ~\cite{huang99}) are:

\begin{itemize}
\item The {\em input field} maps sample locations (e.g.\ on a uniform
grid) to perturbation levels, representing allowable imprecision
before a point becomes indistinguishable from an eigenvalue
(computational details are discussed later).
\item {\em Aggregate} points in a {\em neighborhood graph} (e.g.\
Delaunay triangulation or regular grid), localizing computation to
spatially proximate points.
\item {\em Interpolate} values at new locations from
values at neighboring samples, in this case determining locations of
points with perturbation level belonging to a discrete set (e.g.\
$10^{-1}$, $10^{-2}$, $\ldots$).
\item {\em Classify} neighboring similar-enough objects into
equivalence classes with an {\em equivalence predicate}, in this case
testing equality of field value.\footnote{While an implementation such
as marching squares might combine interpolation,
aggregation, and classification, we view them as conceptually distinct
operations.}
\item {\em Redescribe} each equivalence class of lower-level objects
as a single higher-level object, in this case abstracting connected
points into curves.  The curve can be represented more compactly
and abstractly (e.g.\ with a spline) than its set of sample points.
\end{itemize}

As a consequence of redescription, the next aggregation level can
treat curves as first-class objects, and aggregate, classify, and
redescribe them, for example to find curves nested around a single
eigenvalue.  This higher-level process uses the same operators, but
with different parameters specifying locality, equivalence, and
abstraction.  Ambiguity arises when, for example, not enough sample
points are available to be confident in a curve's location, or in the
separation of two curves.  Ambiguity-directed sampling then optimizes
selection of new locations (e.g.\ near the ambiguity) for data
collection in order to clarify the decision-making.

As this example illustrates, SAL and ambiguity-directed sampling
provide a suitable {\em vocabulary} (e.g.\ distance and similarity
metrics) and {\em mechanism} (bottom-up aggregation and top-down
sampling) to uncover multi-level structures in spatial data sets.
Successful applications include decentralized control
design~\cite{bailey-kellogg99,bailey-kellogg01} weather data
analysis~\cite{huang99}, analysis of diffusion-reaction
morphogenesis~\cite{ordonez00}, and identification of pockets in
$n$-dimensional space and decomposition of a field based on control
influences~\cite{bailey-kellogg-naren01}.

\section{Qualitative Analysis of Correspondence}

Our correspondence mechanism builds on the relationship between
lower-level and higher-level objects in a SAL hierarchy (refer again
to Fig.~\ref{fig:sal}).  The mechanism has two key steps: (1)
establish {\em analogy} as a relation among lower-level constituents
of higher-level objects; (2) establish {\em correspondence} between
higher-level objects as an {\em abstraction} of the analogy between
their constituents.  For example, in object recognition, analogy might
match image and template features, and correspondence might abstract
the analogy as a rigid-body transformation.  Similarly, in level curve
analysis, analogy might match sample points on neighboring curves by
location and local curvature, and correspondence might abstract the
match as a parameterized deformation of spline representations of the
curves.  The analogy between constituents is well-defined only because
of the context of the higher-level objects; higher-level
correspondence then captures a more global view of the local matches.

Tab.~\ref{tab:ops} outlines our correspondence mechanism.  Traditional
SAL operators collect and abstract groups of lower-level objects into
higher-level objects, and establish pairs of higher-level objects for
which correspondence is to be considered.  Two pieces of domain
knowledge are then applied:

\begin{description}
\item[Analogy predicate] indicates pairs (and confidence in the pairs) 
of lower-level objects that are analogous with respect to the
higher-level objects they comprise.  Examples include testing feature
values for each pair of lower-level objects, hashing indices in the
higher-level objects' local coordinate systems, or explicitly
constructing a spatial relation such as a triangulation.  The
predicate can enforce bijective analogy if appropriate.  The {\em
analogize} operator in Tab.~\ref{tab:ops} applies a predicate $a$ for
the constituent objects $l_1 \in h_1$ and $l_2 \in h_2$ of neighboring
(by $g_h$) higher-level object pairs $h_1$ and $h_2$; it returns a
graph labeling the confidence of each analogous object pair.

\item[Correspondence abstraction function] abstracts an analogy
relation on lower-level objects into a description of higher-level
object correspondence.  Examples include redescribing point-wise
matches as a rigid-body transformation or a parameterized deformation.
The abstraction captures confidence in the overall correspondence, for
example computing root-mean squared distance or a Hausdorff metric
between locations or features (e.g.\ local curvature) of analogous
objects.  Correspondence optimization thus entails adjusting the
underlying analogy to maximize the abstracted confidence.  Finally,
while nearby constituents of one object often match nearby
constituents of another, {\em discontinuity} (e.g.\ when an outer
curve envelops two inner ones in Fig.~\ref{fig:compan}) is a
noteworthy event detectable with the abstraction mechanism.  The {\em
correspond} operator in Tab.~\ref{tab:ops} performs correspondence
abstraction from an analogy $g_l$, applying a function $c$ to the
subgraph of $g_l$ on the constituent objects $\ell(h_1)$ and
$\ell(h_2)$ of neighboring (by $g_h$) higher-level objects $h_1$ and
$h_2$; it returns a graph labeling each such higher-level object pair
with the resulting correspondence spatial object.
\end{description}

\begin{table}
\fbox{%
\begin{minipage}{\colwidth}
\begin{enumerate}
\item Given lower-level objects $L$, {\em aggregate}, {\em
classify}, and {\em redescribe} them into higher-level objects $H$.

\item {\em Aggregate} higher-level objects $H$ into a
neighborhood graph $G_H$ localizing potential correspondence.

\item Apply an {\em analogy predicate} to relate constituent
lower-level objects of neighboring higher-level objects. \\
{\small
{\it analogize} $: G_H \times (L \times L \rightarrow \R)
\rightarrow G_L$\\
$(g_h,a) \mapsto \set{\set{l_1,l_2,a(l_1,l_2)}\,|\,\set{h_1,h_2} \in g_h, 
l_1 \in h_1, l_2 \in h_2}$
}
where $l_i \in h_i$ represents constituency.

\item Apply a {\em correspondence abstraction function}
to establish correspondence between higher-level objects based
on the analogy on their constituent lower-level objects.
{\small
{\it correspond} $: G_H \times G_L \times (G_L \rightarrow SO)
\rightarrow G_H'$ \\
$(g_h,g_l,c) \mapsto \set{\set{h_1,h_2,c(g_l[\ell(h_1) \cup
\ell(h_2)])}\,|\, \set{h_1,h_2} \in g_h}$
}
where $\ell(\cdot)$ obtains constituents, $g_l[\cdot]$ is the
subgraph for the given nodes, and $c$ returns a spatial object ($SO$)
representing the abstracted correspondence.
\end{enumerate}
\end{minipage}
}
\vspace*{-0.5\baselineskip}
\caption{Qualitative correspondence analysis mechanism, including
formal definitions of new operators.}
\label{tab:ops}
\end{table}

An aggregation/correspondence hierarchy establishes a distribution of
possible high-level models for an input instance, thereby posing a
{\em model selection problem}: choose the one that (e.g.\ in a
maximum-likelihood sense) best matches the data.  Our mechanism
supports model selection in two key ways. (1) The operators estimate
and optimize confidence in correspondence.  Since correspondence
implies mutual support among parts of a model, it can allow relatively
high-confidence model selection even with sparse, noisy data.  (2) The
operators bridge the lower-/higher-level gap.  This allows weaknesses
and inconsistencies detected in higher-level correspondence to focus
lower-level data collection to be maximally effective for model
disambiguation.

\section{Applications in Experimental Algorithmics}

We present two case studies applying our analysis framework to
experimental algorithmics problems.  For each we describe the
underlying numerical analysis problem, our particular solution
approach, and results.  To the best of our knowledge, these are the
{\em first} systematic algorithms for performing complete imagistic
analyses (as opposed to relying on human visual
inspection~\cite{precise}), and which focus data collection and
evaluate models until a high-confidence model is obtained.

\subsection{Matrix Spectral Portrait Analysis}

Our first case study focuses on the previously introduced task of
matrix spectral portrait analysis (Fig.~\ref{fig:compan}).  Formally,
the spectral portrait of a matrix $\mathcal{A}$ is defined as:
\begin{equation}\label{eq:portrait}
{\mathcal P}(z) = \log {\parallel \mathcal{A} \parallel}_2 \,\, 
{\parallel {(\mathcal{A} - zI)}^{-1}\parallel}_2,
\end{equation}
where $I$ is the identity matrix.  The singularities of this map are
located at the eigenvalues of the matrix, and the analysis determines
the sensitivity of computation to numerical imprecision by analyzing
how the map decreases moving away from the eigenvalues.  As discussed
in the introduction, the region enclosed by a level curve of a certain
magnitude contains all points that act as ``equivalent'' eigenvalues
under perturbations of that magnitude.  For example,
Fig.~\ref{fig:compan} illustrates that, under large enough
perturbation, eigenvalues at 3 and 4 are indistinguishable; after
larger perturbations, their joint contour also surrounds the
eigenvalue at 2, and then the eigenvalue at 1.  This illustrates that
sensitivity analysis is solved by identifying values at which level
curves merge.

Tab.~\ref{tab:ellipse-code} describes qualitative correspondence
analysis of spectral portraits.  Data are collected by computing
Eq.~\ref{eq:portrait}; the analysis determines perturbation
equivalence of eigenvalues by detecting curve merges via level curve
correspondence.  The first aggregation level generates samples on a
coarse regular grid around the known eigenvalue locations, and then
interpolates and abstracts level curves as in the spatial aggregation
example.  The second aggregation level finds correspondence among
these curves from a Delaunay triangulation analogy of their
constituent points.  It tracks correspondence outward from the
eigenvalues to establish a model of merge events, establishing
confidence in such a model by good analogy (most points match) and
good curve separation (regularly-spaced samples separate the two
curves, providing evidence that contours do not merge at a smaller
perturbation).  Ambiguity-directed sampling generates additional data
in order to separate curves and to ensure that each eigenvalue pair is
merged at some perturbation.

\begin{table}[t]
\fbox{%
\begin{minipage}{\colwidth}
{\bf Input}: matrix ${\mathcal A}$, eigenvalues $E$,
perturbation levels $V$. \\
{\bf Output}: $\set{(E_i,E_j,v_{ij})}$ such that eigenvalues $E_i$ and
$E_j$ are equivalent with respect to perturbation of $v_{ij} \in V$.

\medskip

{\bf Level one}:
\begin{closeitemize}
\item Data collection: Eq.~\ref{eq:portrait}.
\item Initial samples $P$: points on coarse regular grid.
\item Output: level curves $C$.
\item Aggregation: aggregate grid; interpolate points $I$ at
values in $V$; classify by perturbation; redescribe into curves.
\item Aggregation 2: $G_I$ = triangulation of points $I$.
\end{closeitemize}

\medskip

{\bf Level two}:
\begin{closeitemize}
\item Input: curves $C$.
\item Output: problem output $\set{(E_i,E_j,v_{ij})}$.
\item Aggregation: $(C_k,C_l) \in G_C$ iff constituent points are
neighbors in $G_I$.
\item Correspondence:
  \begin{closeitemize}
  \item Analogy: cross-curve neighbors in $G_I$.
  \item Abstraction: $C_k,C_l \mapsto (m_k,m_l,\theta_{kl})$, for
$m_i$\% and $m_j$\% constituent points matched, and samples $P'
\subseteq P$ between $C_k$ and $C_l$ separated by no more than
$\theta_{kl}$ in angle around the enclosed eigenvalue.
  \end{closeitemize}
\item Model evaluation: follow correspondence outward from each
pair of eigenvalues $(i,j)$; evaluate confidence with respect to
$(m_k,m_l,\theta_{kl})$.
\item Sampling: 
  \begin{closeitemize}
  \item When no $(E_i,E_j,v_{ij})$ for some $(i,j)$, expand grid.
  \item When some $\theta_{kl}$ too large, subsample on finer grid.
  \end{closeitemize}
\end{closeitemize}
\end{minipage}
}
\vspace*{-0.5\baselineskip}
\caption{Correspondence mechanism instantiation for spectral portrait
analysis.}
\label{tab:ellipse-code}
\end{table}

\begin{figure}
\begin{center}
\begin{tabular}{cc}
\includegraphics[width=3in]{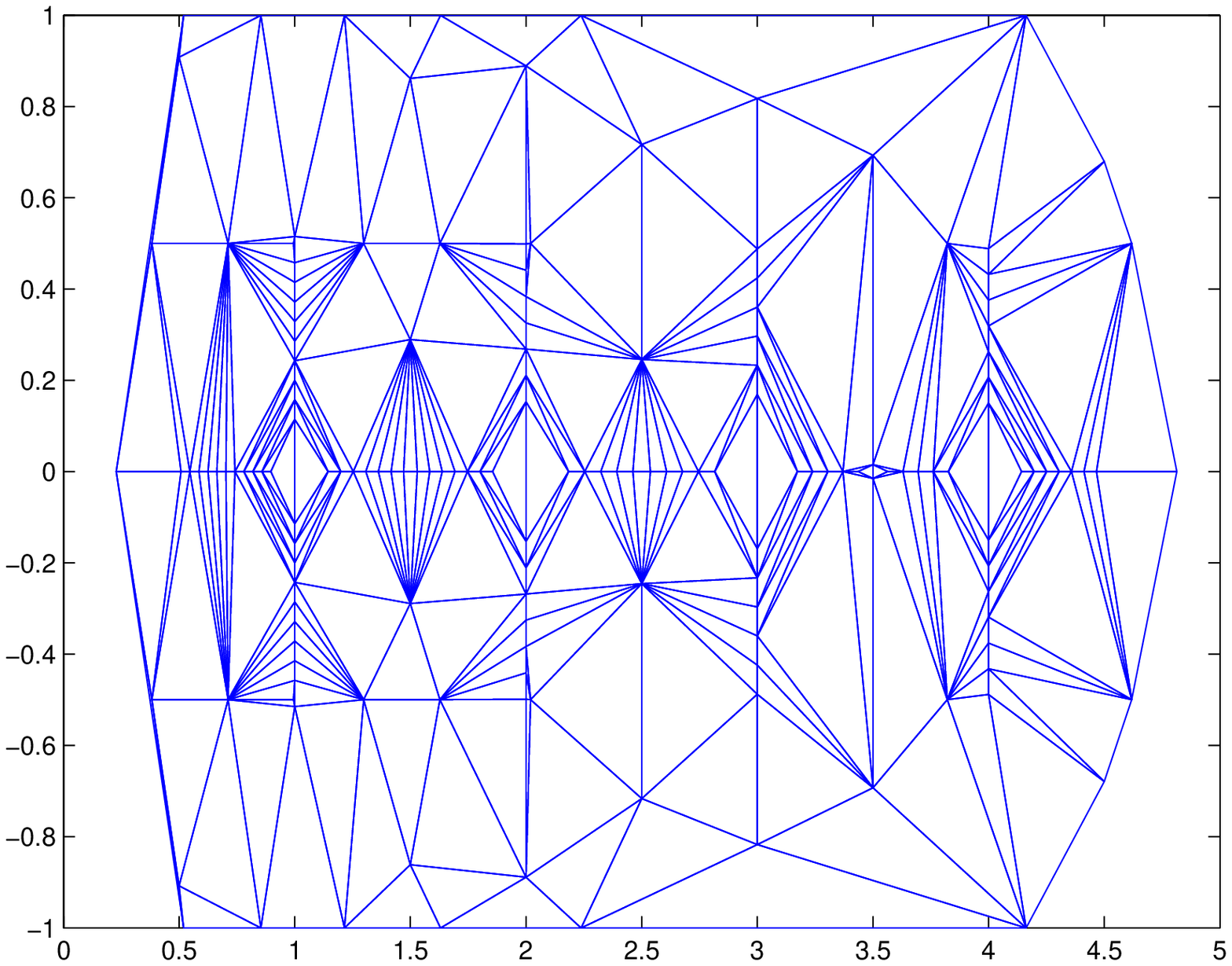} & 
\includegraphics[width=3in]{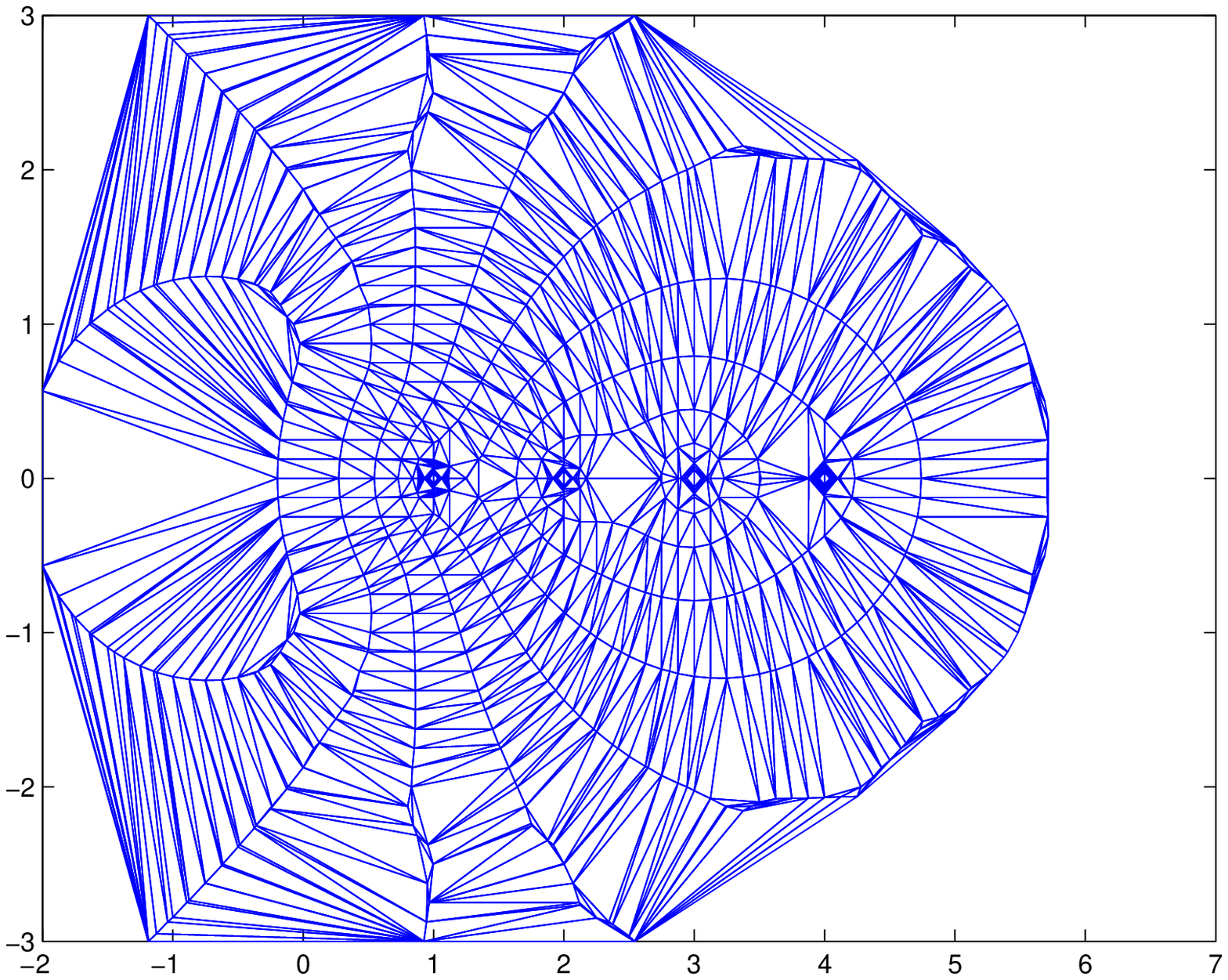} \\
\includegraphics[width=3in]{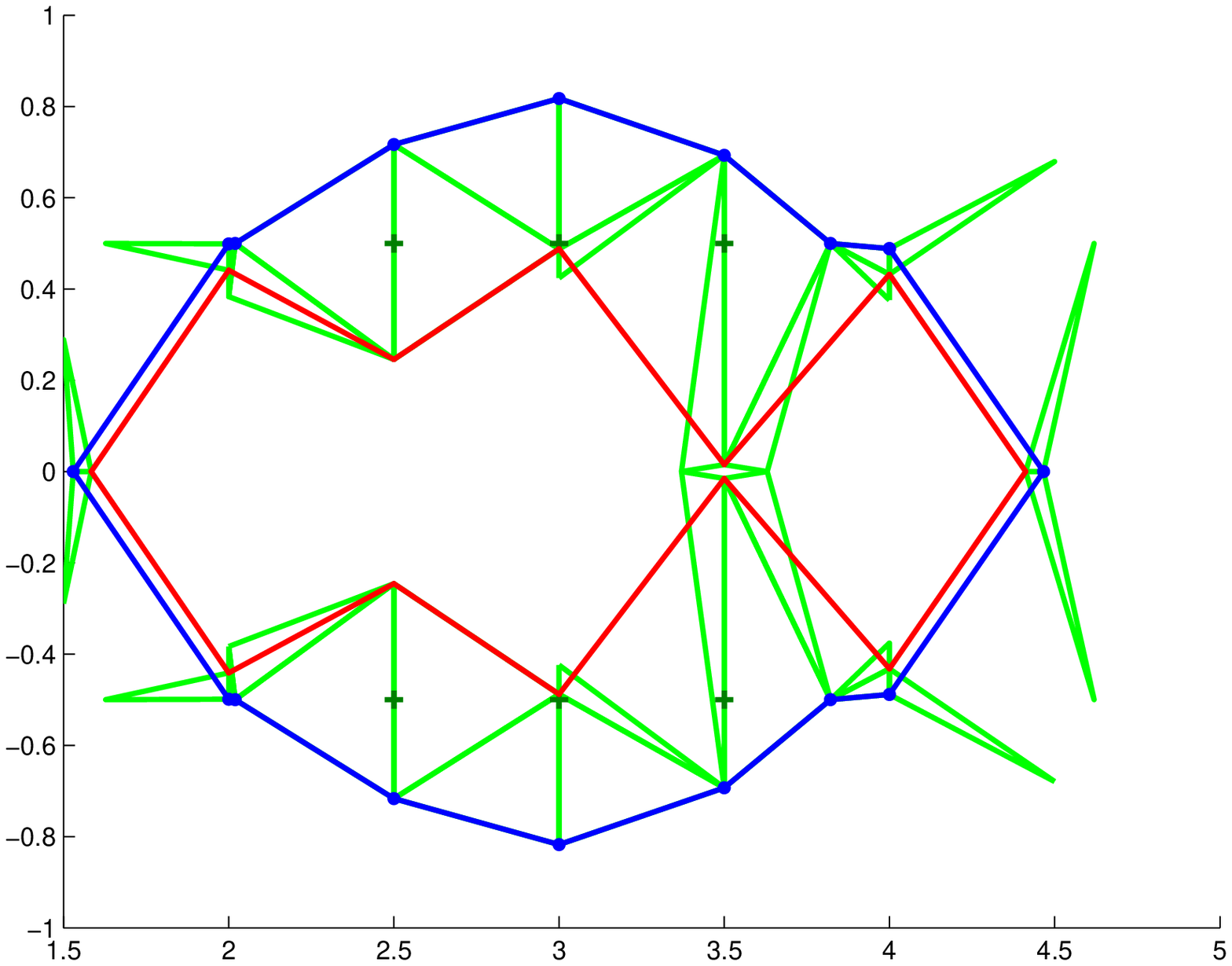} &
\includegraphics[width=3in]{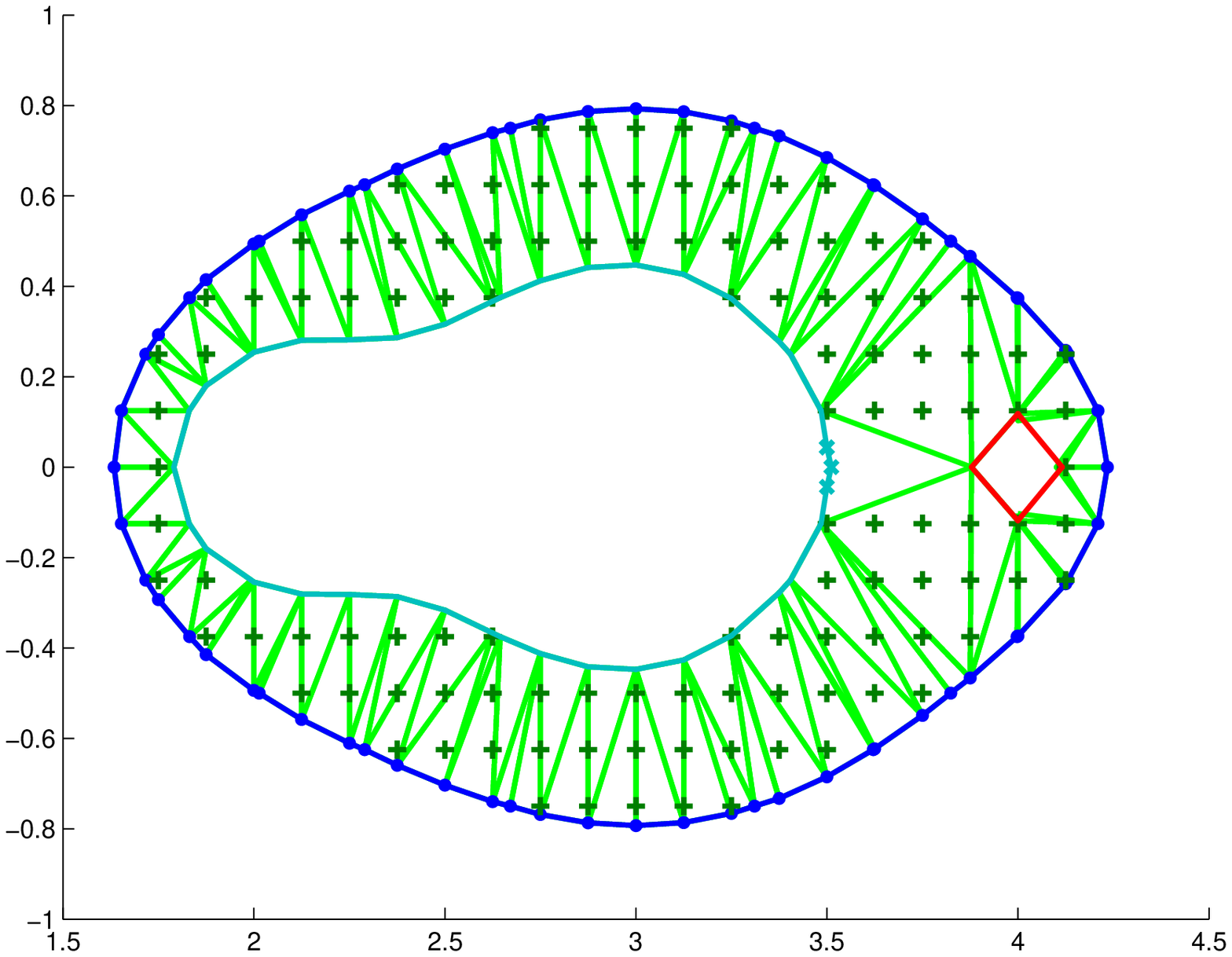} \\
\includegraphics[width=2.5in]{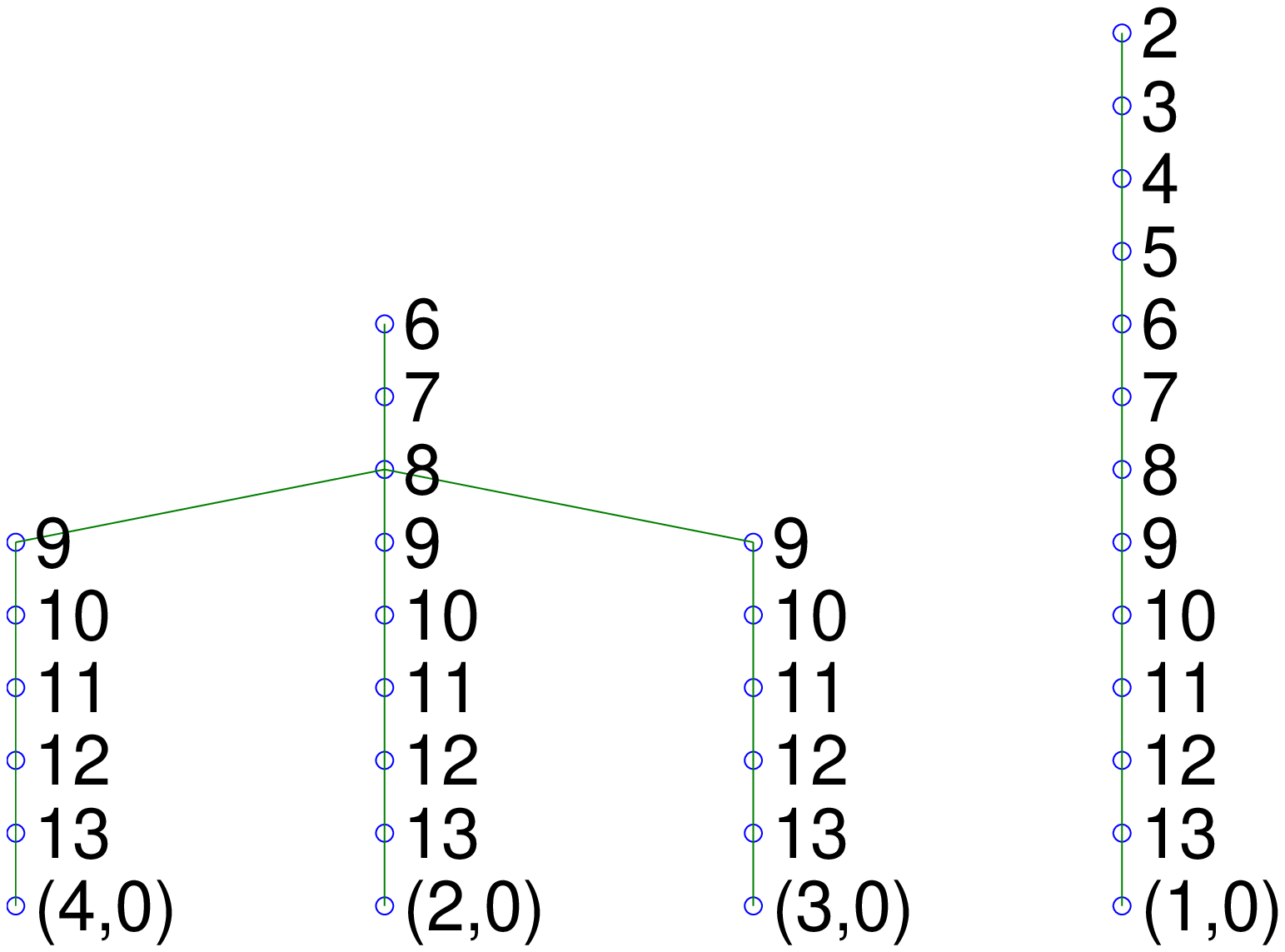} &
\includegraphics[width=2.5in]{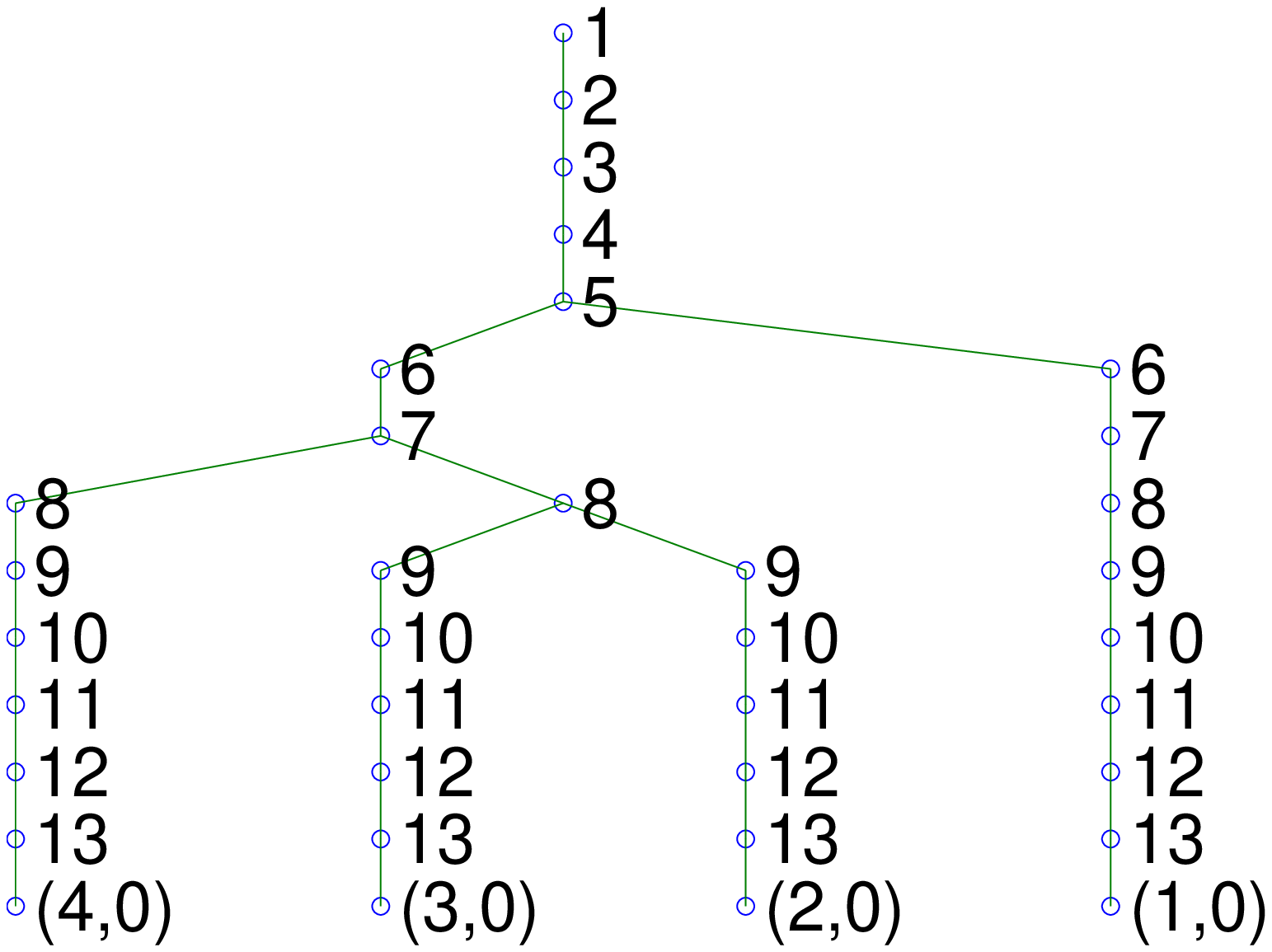}
\end{tabular}
\end{center}
\caption{Example of correspondence analysis of spectral
portraits for (left) a small, coarse grid, and (right) an extended,
subsampled grid.  (top) Delaunay triangulation analogy for
interpolated points comprising contours.  (middle) Analogy at an
example merge event; separating samples marked with ``+''.  (bottom)
Curve-tracking tree: eigenvalues at bottom; node for curves labeled
with perturbation level.  Merge events indicate at what perturbation
level descendant eigenvalues are indistinguishable.}
\label{fig:ellipse-demo}
\end{figure}

Fig.~\ref{fig:ellipse-demo} demonstrates the application of the
mechanism to the companion matrix of the polynomial $(x-1)^3 (x-2)^3
(x-3)^3 (x-4)$~\cite{precise} (see also Fig.~\ref{fig:compan}).  The
initial grid has at least one sample between each eigenvalue (in the
example, a resolution of 0.5) and extends one unit beyond the bounding
box of the eigenvalues.  While correspondence is found even with this
coarse grid (left of Fig.~\ref{fig:ellipse-demo}), the confidence is
not high (only a few samples separate the curves), and no curve
surrounds all eigenvalues.  Ambiguity-directed sampling computes
additional points on a finer, larger grid (right of
Fig.~\ref{fig:ellipse-demo}), yielding the correct results.  We have
applied this same mechanism to a variety of polynomial companion
matrices with different numbers and spacings of roots; in each case,
the correspondence mechanism correctly identifies the correct model
with high confidence after 1-3 subsamples and 1-3 grid expansions.

\subsection{Qualitative Computation of Jordan Forms}

Our second case study focuses on analysis of the {\em Jordan
decomposition} of a matrix.  Matrix decomposition is an important
technique, revealing pertinent features of a matrix and supporting
algorithmic techniques in areas including data analysis, PDEs, and
linear algebra.  The Jordan decomposition reveals the eigenstructure
of a matrix as follows.  Consider a matrix $\mathcal{A}$ that has $r$
independent eigenvectors with eigenvalues $\lambda_i$ of multiplicity
$\rho_i$.  The Jordan decomposition of $\mathcal{A}$ contains $r$
upper triangular ``blocks,'' as revealed by the diagonalization:
{\small
\[
\mathcal{B}^{-1} \mathcal{A} \mathcal{B} = 
\left[ \begin{array}{cccc}
       {\mathcal{J}}_1 & & & \\
       & {\mathcal{J}}_2 & & \\
       & & \cdot & \\
       & & & {\mathcal{J}}_r\\
       \end{array} \right],
\,\,
{\mathcal{J}}_i = 
\left[ \begin{array}{cccc}
       \lambda_i & 1 & & \\
       & \lambda_i & 1 & \\
       & & \cdot & 1 \\
       & & & \lambda_i \\
       \end{array} \right]
\]
}

\begin{figure}
\begin{center}
\begin{tabular}{cc}
\includegraphics[width=3in]{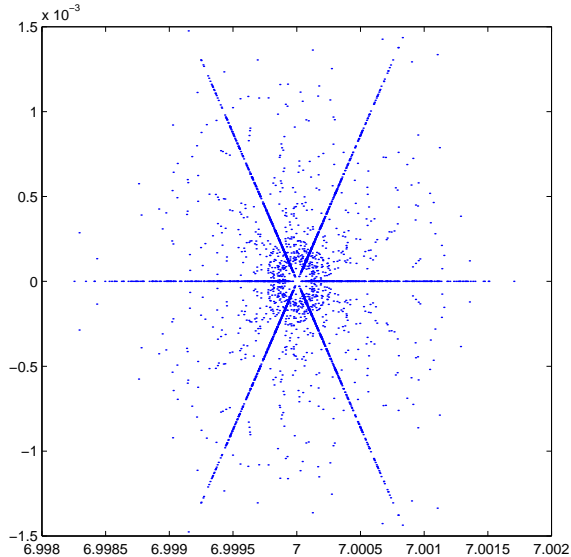}
\end{tabular}
\end{center}
\caption{Superimposed spectra for assessing the Jordan form
of the Brunet matrix.}
\label{fig:jordan}
\end{figure}

The typical approach to computing the Jordan form leads to a
numerically unstable algorithm~\cite{golub-loan}; taking extra care
usually requires more work than the original computation!  Recently,
however, an experimental algorithmics approach, inferring multiplicity
from a geometric analysis of eigenvalue perturbations, has proved
successful~\cite{precise}.  It is well known that the {\it
computed eigenvalues} corresponding to the actual value $\lambda_i$
are given by:
\begin{equation}\label{eq:jordan-perturb}
\lambda_i + |\delta|^{1\over{\rho_i}} e^{{i\phi}\over{\rho_i}},
\end{equation}
where $\lambda_i$ is of multiplicity $\rho_i$, and the phase $\phi$ of
the perturbation $\delta$ ranges over
\set{$2\pi, 4\pi, \ldots, 2\rho_i \pi$} if $\delta$ is positive and
over \set{$3\pi, 5\pi, \ldots, 2(\rho_i+1) \pi$} if $\delta$ is
negative.  The insight of~\cite{precise} is to graphically {\it
superimpose} numerous such perturbed calculations so that the
aggregate picture reveals eigenvalue multiplicity.  The phase
variations imply that computed eigenvalues lie on the vertices of a
regular polygon with $2 \rho_i$ sides, centered on $\lambda_i$, and
with diameter influenced by $|\delta|$.  For example,
Fig.~\ref{fig:jordan} shows perturbations for the 8-by-8 Brunet matrix
with Jordan structure $(-1)^1 (-2)^1 (7)^3 (7)^3$~\cite{precise}, for
$\delta \in [2^{-50},2^{-40}]$.  The six ``sticks'' around the
eigenvalue at 7 clearly reveal that its Jordan block is of size
3.\footnote{The multiplicity of the second eigenvalue at 7 is revealed
at a smaller perturbation level.}  The ``noise'' in
Fig.~\ref{fig:jordan} is a consequence of having two Jordan blocks
with the same eigenvalue and size, and a ``ring'' phenomenon studied
in~\cite{edelman-ma}; we do not attempt to capture these effects in
this paper.

\begin{figure}
\begin{center}
\includegraphics[width=3in]{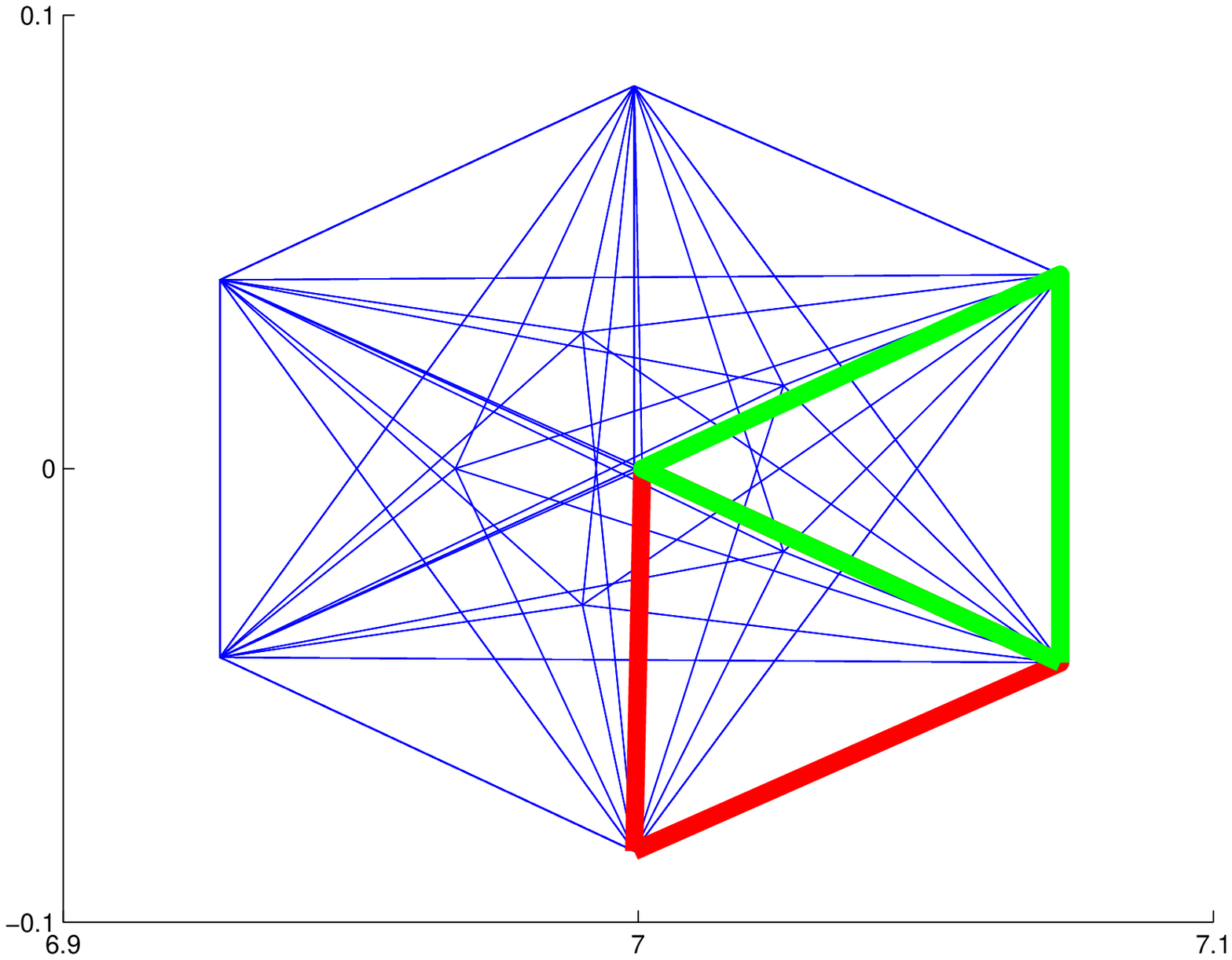}\hspace*{0.1in}\includegraphics[width=3in]{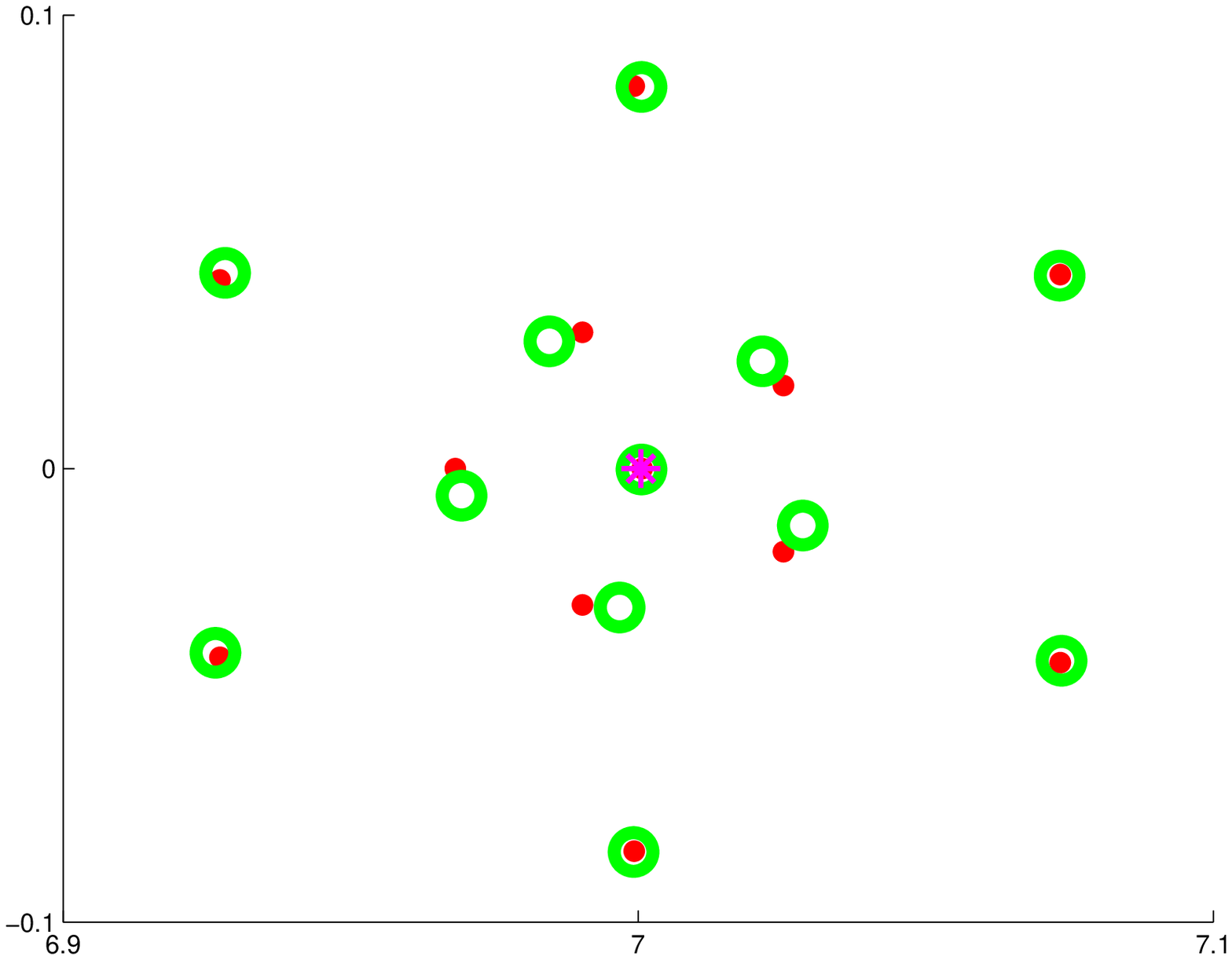}
\\
\vspace*{0.1in}
\includegraphics[width=3in]{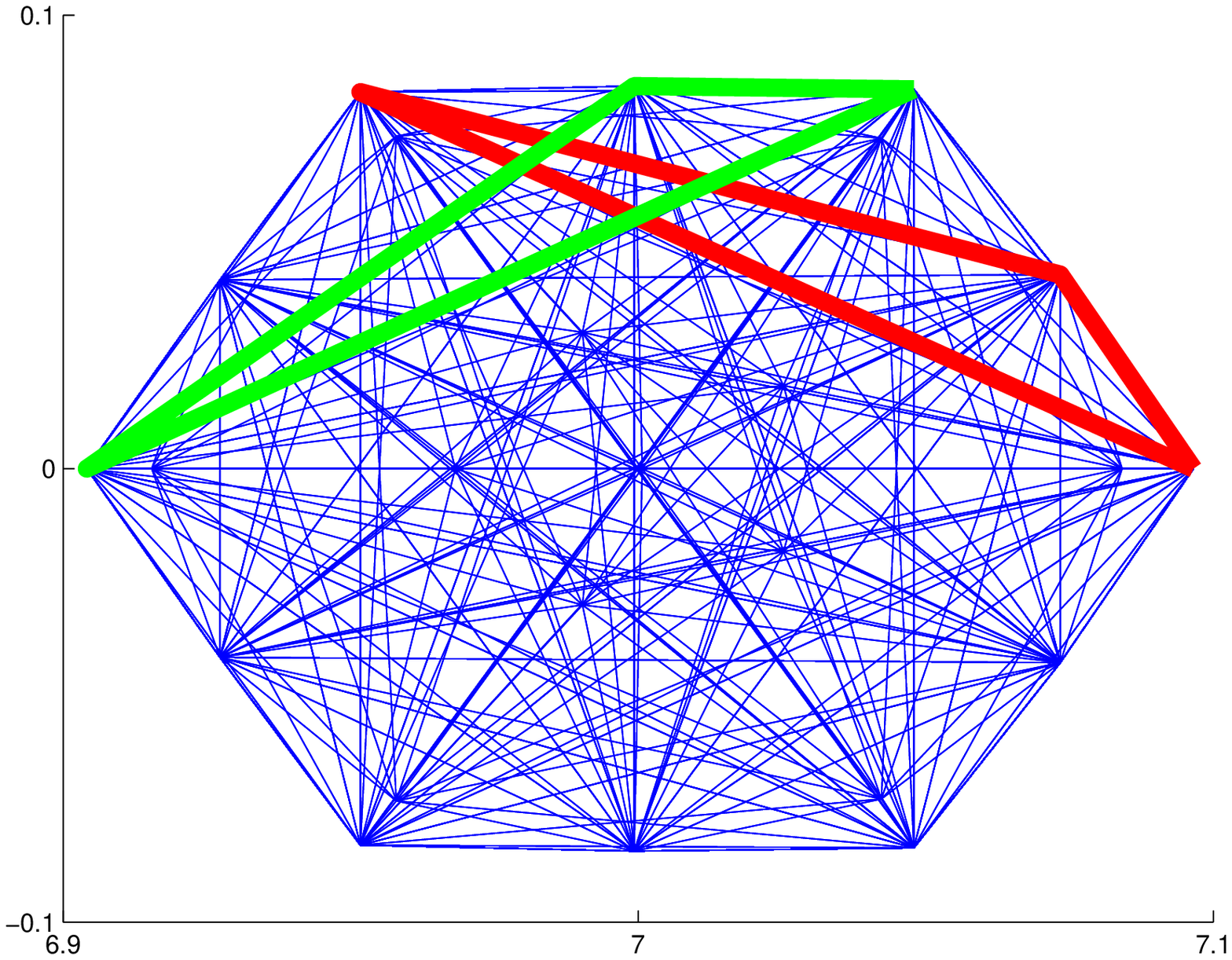}\hspace*{0.1in}\includegraphics[width=3in]{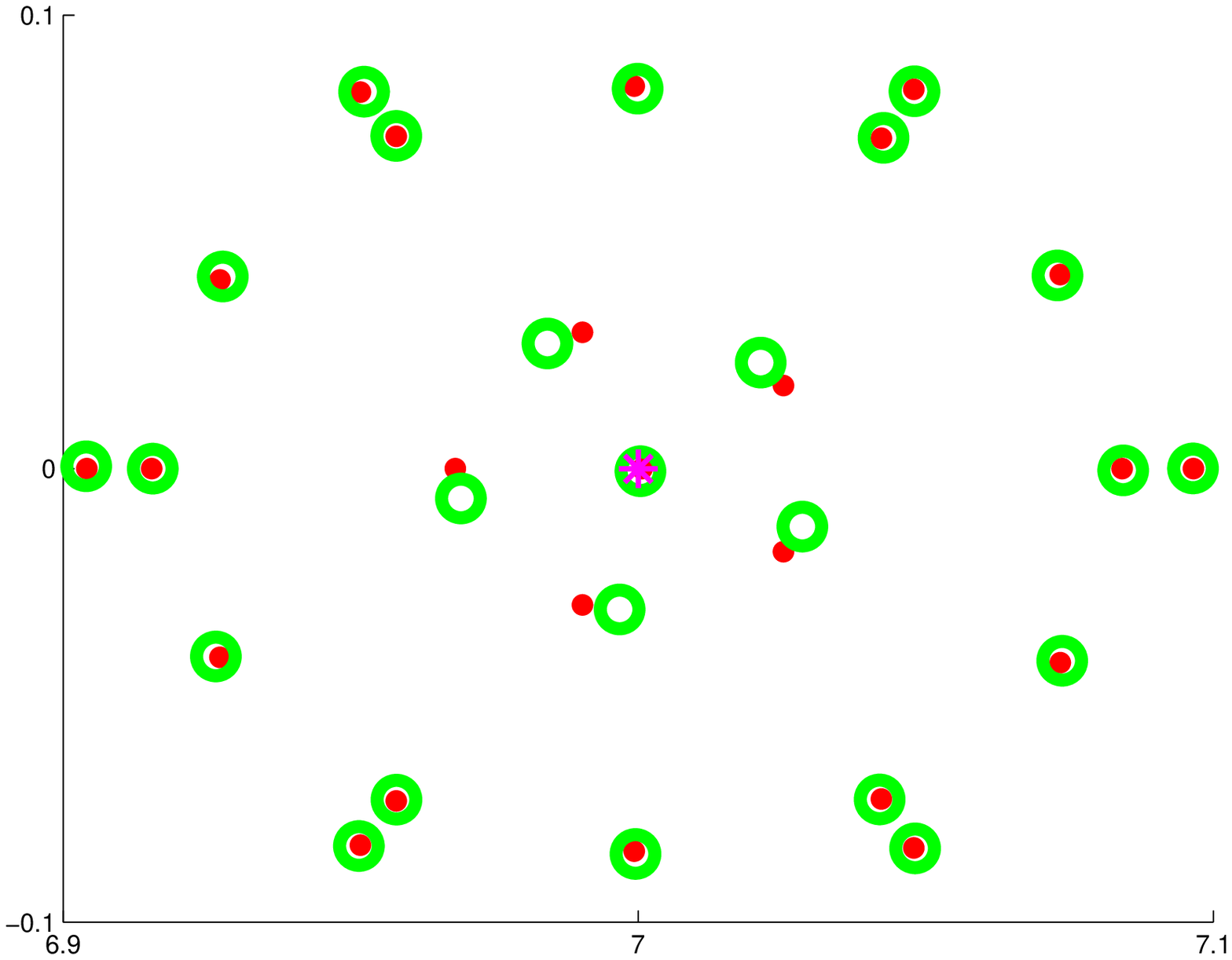}
\\
\vspace*{0.1in}
\includegraphics[width=3in]{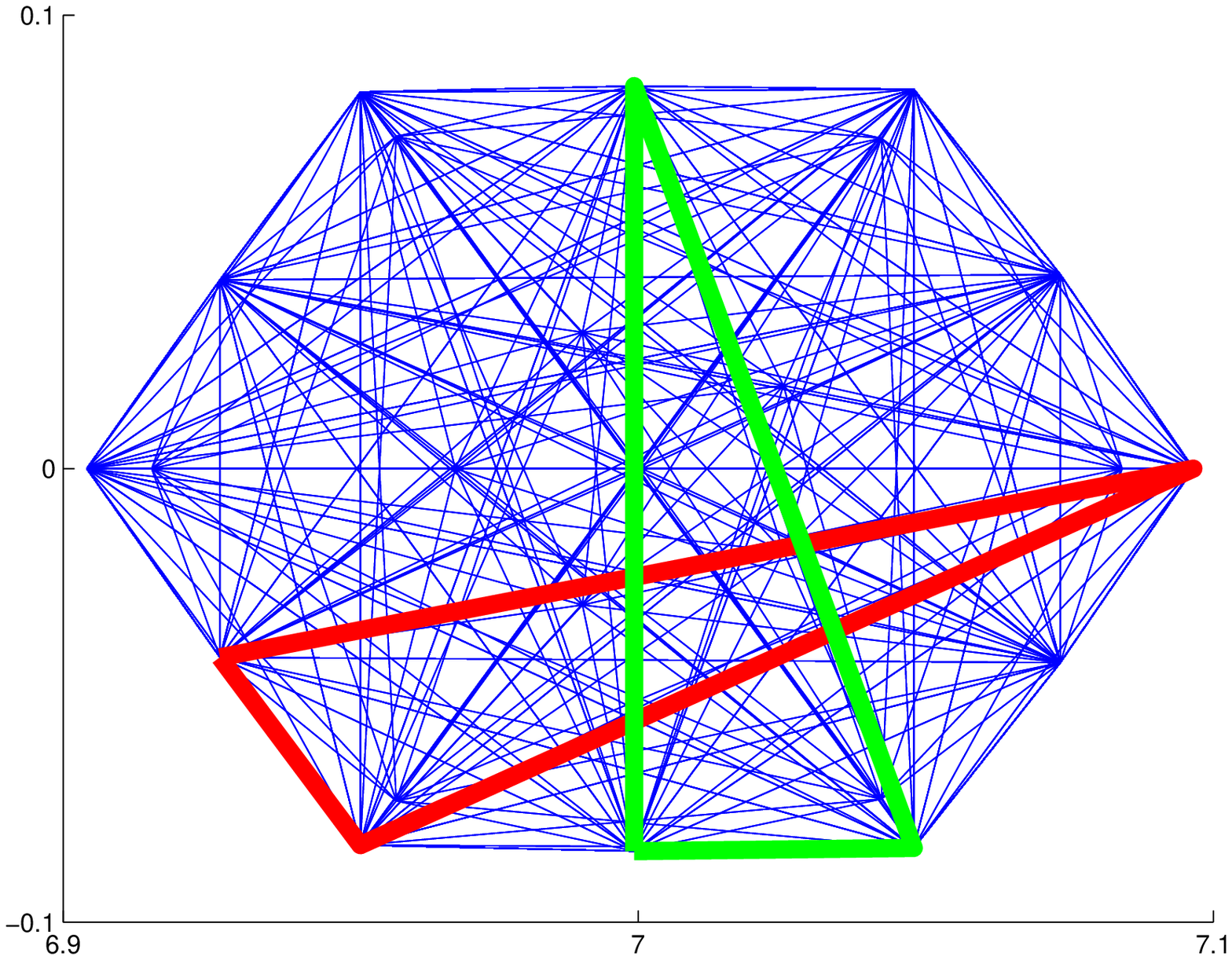}\hspace*{0.1in}\includegraphics[width=3in]{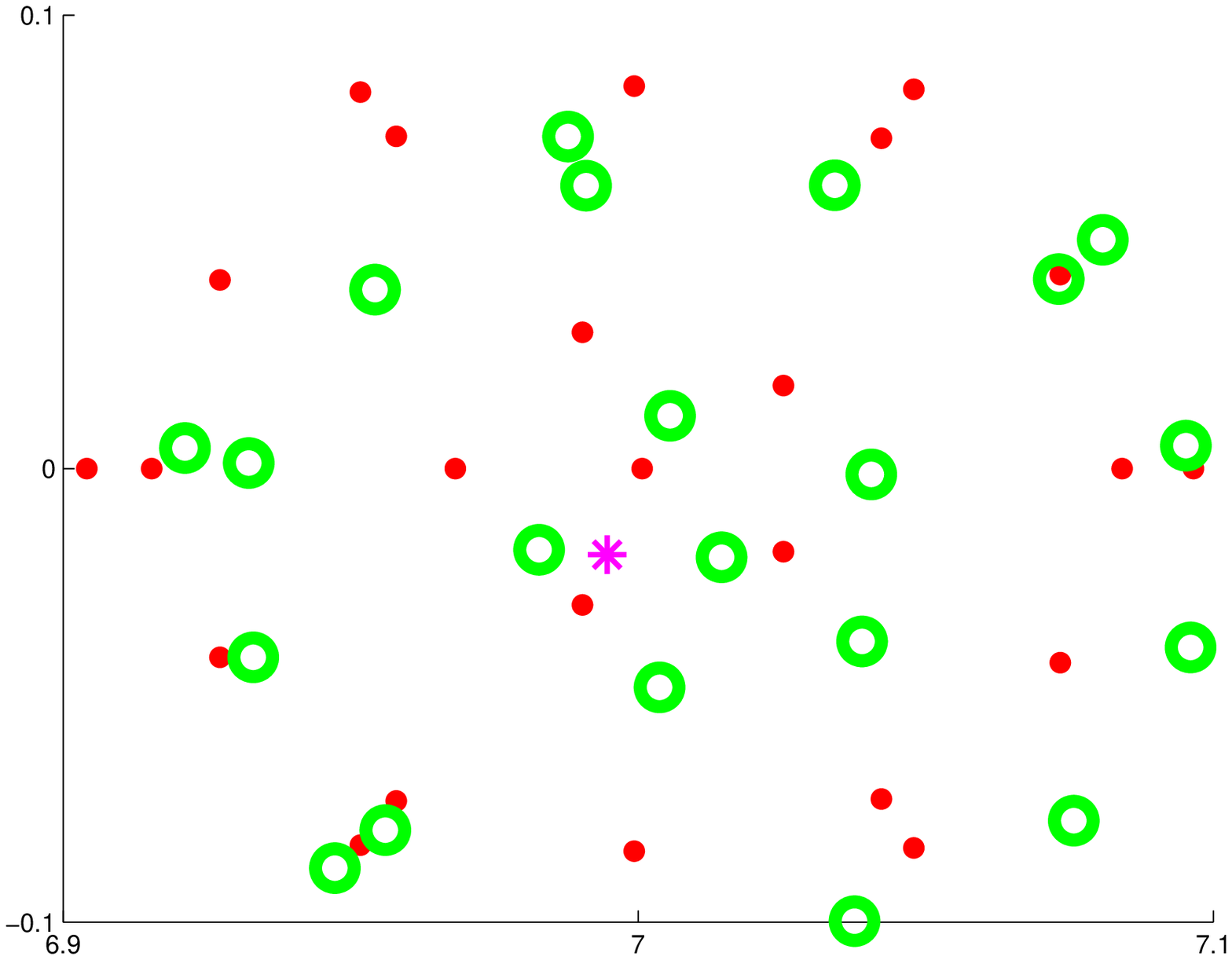}
\end{center}
\caption{Example of correspondence analysis of Jordan form, for (top)
small sample set; (middle) larger sample set; (bottom) larger sample
set but lower-scoring model.  (left) Approximately-congruent
triangles.  (right) Evaluation of correspondence in terms of match
between original (red dots) and rotated (green circles) samples.}
\label{fig:star-demo}
\end{figure}

Tab.~\ref{tab:star-code} describes qualitative correspondence analysis
of Jordan form.  Data are collected by randomly perturbing at a
specified magnitude $\delta$; the analysis determines multiplicity by
detecting {\em symmetry} correspondence in the samples.  The first
aggregation level collects the samples for a given $\delta$ into
triangles.  The second aggregation level finds congruent triangles via
geometric hashing~\cite{hash}, and uses congruence to establish
analogy among triangle vertices.  Correspondence abstracts the analogy
into a rotation about a point (the eigenvalue), and evaluates whether
each point rotates onto another and whether matches define regular
polygons.  Ambiguity-directed sampling collects additional random
samples as necessary.  A third level then compares rotations across
different perturbations, re-visiting perturbations or choosing new
perturbations in order to disambiguate.  Fig.~\ref{fig:star-demo}
demonstrates this mechanism on the Brunet matrix discussed above.  The
top part uses a small set of sample points, while the bottom two parts
use a larger set and illustrate a good vs.\ bad correspondence.

We organized data collection into rounds of 6-8 samples each and
experimented with three policies on where next to collect data after
completing a round: (1) at the same perturbation level, (2) at a
higher perturbation level, or (3) at the same perturbation level
unless the number of posited models increased (model
``hallucination,'' as in bottom of Fig.~\ref{fig:star-demo}).  We tested
10 matrices across 4-10 perturbation levels each, as described
in~\cite{precise}.  We varied a tolerance parameter for triangle
congruence from 0.1 to 0.5 (effectively increasing the number of
models posited) and determined the number of rounds needed to
determine the Jordan form.  Policy 1 required an average of 1 round at
a tolerance of 0.1, up to 2.7 rounds at 0.5.  Even with a large number
of models proposed, additional data quickly weeded out bad models.
Policy 2 fared better only for cases where policy 1 was focused on
lower perturbation levels, and policy 3 was preferable only for the
Brunet-type matrices.  In other words, there is no real advantage to
moving across perturbation levels!  In retrospect, this is not
surprising since our Jordan form computation treats multiple
perturbations (irresp.\ of level) as independent estimates of
eigenstructure.

\begin{table}[t]
\fbox{%
\begin{minipage}{\colwidth}
{\bf Input}: matrix ${\mathcal A}$, perturbations $\set{\delta_1,
\ldots, \delta_m}$, region $R$. \\
{\bf Output}: eigenvalue $\lambda$ and multiplicity $\rho$ in region
$R$.

\medskip
{\bf Level one}:
\begin{closeitemize}
\item Data collection: for $\delta_i$, compute random normwise
perturbation of ${\mathcal A}$ as $a_{ij} \pm 2^{(1 - \delta_i)}
{\parallel \mathcal{A} \parallel}_{\infty}$, yielding $P_i$.
\item Initial samples: At some level $i$.
\item Output: triangles $T_i$.
\item Aggregation: triangulate $P_i$ (for efficiency, require 2 vertices 
on convex hull).
\end{closeitemize}

\medskip
{\bf Level two}:
\begin{closeitemize}
\item Input: Triangles $T_i$.
\item Output: set of rotations $(x,y,\theta)$.
\item Aggregation: congruent triangles by geom.\ hashing.
\item Correspondence:
  \begin{closeitemize}
  \item Analogy: by triangle congruence.
  \item Abstraction: $(t_j,t_k) \mapsto (x,y,\theta,d,r)$,
where rotation $(x,y,\theta)$ yields same analogy, with
RMSD $d$ between $P_i$ and rotated, and polygon regularity $r
= \parallel \parallel p_a-p_b \parallel - \parallel p_b-p_c \parallel
\parallel$, where rotation maps $p_a \rightarrow p_b \rightarrow p_c$.
  \end{closeitemize}
\item Model evaluation: confidence with respect to $d$ and $r$; priors
support rotations around $(x,y)$ in convex hull of $P_i$ and by
$\theta$ corresponding to ``reasonable'' multiplicity.
\item Sampling: for multiple ``good'' models, collect additional
random samples at perturbation $\delta_i$.
\end{closeitemize}

\medskip
{\bf Level three}:
\begin{closeitemize}
\item Input: $\set{(x_{ij},y_{ij},\theta_{ij})}$ over models ($j$)
from chosen perturbation levels ($i$).
\item Output: $(\lambda,\rho)$.
\item Aggregation: clustering in $(x,y,\theta)$-space.
\item Model evaluation: for $\lambda=(x,y),\rho=\pi/\theta$, take
joint probability over $i,j$ of $(x_{ij},y_{ij},\theta_{ij}) \approx
(x,y,\theta)$.
\item Sampling: for high entropy in model evaluation, add samples and
re-evaluate at outlier $\delta_j$ or try new $\delta_k$.
\end{closeitemize}
\end{minipage}
}
\vspace*{-0.5\baselineskip}
\caption{Correspondence mechanism instantiation for Jordan form
analysis.}
\label{tab:star-code}
\end{table}

\section{Discussion}

Our general mechanism for uncovering and utilizing correspondence has
proved successful on challenging problems from experimental
algorithmics.  The mechanism leverages properties such as locality,
continuity, and decomposability, which are exhibited by the
applications studied here as well as many physical systems.
Decomposability and locality allow us to qualify correspondence in a
manner that drives data collection.  Contintuity allows correspondence
to obtain confidence in a model even with sparse, noisy data:
consistent matches among nearby constituents mutually reinforce each
other, allowing correspondence abstraction to detect and filter out
inconsistent interpretations.

Correspondence can be an appropriate analysis tool for a variety of
reasons.  In the spectral portrait application, level curves summarize
groups of eigenvalue computations, so their high-level correspondence
aids characterization of the underlying computation.  In the Jordan
form application, however, the higher-level entity has no significance
as a geometric object in numerical analysis terms, but correspondence
is applicable due to the semantics of superposition.  This semantics
also leads to phenomena such as hallucination (given enough samples,
any pattern can be found), requiring a more careful treatment of
decomposition.

Our work is similar in spirit to that of~\cite{huang99} for weather
data interpretation, and can be seen as a significant generalization,
formalization, and application of techniques studied there for finding
correspondence in meteorological data.  Similarly, our correspondence
framework captures and generalizes the computation required in object
recognition, allowing the body of research developed there to be
applied to a broader class of applications, such as experimental
algorithmics.  Compared to traditional manual analyses of graphical
representations in experimental algorithmics, the algorithmic nature
of our approach yields advantages such as model evaluation and
targeted sampling.  As with compositional modeling~\cite{compo}, we
advocate targeted use of domain knowledge, and as with
qualitative/quantitative model selection (e.g.~\cite{ironi}), we seek
to determine high level models for empirical data.  Our focus is on
problems requiring particular forms of domain knowledge to overcome
sparsity and noise in spatial datasets. A possible direction of future
work is to explore if the inclusion-exclusion methodology popular in
grid algorithms~\cite{gallopoulos-portrait} is also useful for
tracking correspondence.

Our long-term goal is to study data collection policies and their
relationships to qualitative model determination.  The notion of
estimating problem-solving performance by collecting data (and vice
versa) is reminiscent of reinforcement learning~\cite{boyan00} and
active learning~\cite{jordan-jair}.  The decomposable nature of SAL
computations promises to (i) support the design of efficient,
hierarchical algorithms for model estimation and (ii) provide a deeper
understanding of the recurring roles that correspondence plays in
spatial data analysis.

\bibliographystyle{plain}
\bibliography{corr}

\end{document}